\documentclass[10pt,twocolumn,letterpaper]{article}

\PassOptionsToPackage{table}{xcolor}  
\usepackage[pagenumbers]{cvpr} 

\definecolor{cvprblue}{rgb}{0.21,0.49,0.74}
\usepackage[pagebackref,breaklinks,colorlinks,allcolors=cvprblue]{hyperref}

\usepackage{multirow}
\usepackage{xcolor}
\usepackage{pifont}

\title{WaterClear-GS: Optical-Aware Gaussian Splatting for Underwater Reconstruction and Restoration}

\author{
Xinrui Zhang, Yufeng Wang, Zesheng Wang, Dacheng Qi, Wenrui Ding, Shuangkang Fang\\
Beihang University\\
{\tt\small xr\_zhang@buaa.edu.cn}
}

\begin{document}
\maketitle
\begin{abstract}
Underwater 3D reconstruction and appearance restoration remain challenging due to the complex optical properties of water, such as wavelength-dependent attenuation and scattering. 
Existing Neural Radiance Fields (NeRF)-based approaches often suffer from slow rendering and limited practicality, while vanilla 3D Gaussian Splatting (3DGS) lacks an effective mechanism to account for underwater degradation.
To address this, we propose WaterClear-GS, a physics-informed reformulation of underwater Gaussian splatting that models degradation as intrinsic Gaussian attributes rather than external medium fields.
Furthermore, we adopt a dual-branch optimization strategy that preserves underwater photometric consistency while encouraging a practical separation between restoration-oriented latent clean appearance and degradation.
This strategy is enhanced by depth-guided geometry regularization and perception-driven image supervision, together with exposure constraints, spatially adaptive regularization, and physics-informed spectral regularization, which collectively promote spatial coherence and plausible visual appearance.
Extensive experiments on standard benchmarks and our collected dataset demonstrate that WaterClear-GS achieves strong performance on both novel view synthesis (NVS) and underwater image restoration (UIR) tasks, while maintaining 160+ FPS real-time rendering.
The code will be available at \href{https://sherii888.github.io/WaterClear-GS/}{this https URL}.
\end{abstract}    
\section{Introduction}
\label{sec:intro}

\begin{figure}[!t]
\centering
\includegraphics[width=\columnwidth]{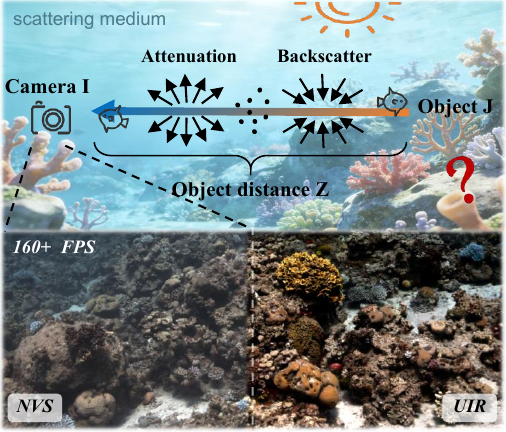}
\caption{Illustration of underwater imaging and our results. \textit{Top}: Light from objects is affected by significant attenuation and scattering, which intensify with distance. \textit{Bottom}: Our method enables high-quality rendering at over 160 FPS and restores plausible clean appearance.}
\label{fig:motivation}
\end{figure}

Underwater 3D reconstruction presents unique opportunities for marine exploration~\cite{marine2,marine}, autonomous underwater navigation~\cite{navigation2,navigation}, and archaeological research~\cite{archaeology,archaeology2}. 
However, the complex optical properties of the underwater environment pose significant challenges. 
In contrast to imaging in clear-air environments, underwater imaging suffers from severe degradation due to wavelength-dependent light attenuation and scattering effects caused by water molecules and suspended particles. These phenomena result in pronounced color distortion, diminished contrast, and suppression of fine structural details, as shown in Figure~\ref{fig:motivation}.
While traditional single-image underwater restoration methods~\cite{whitebalance,initial,underwater} can leverage statistical priors or data-driven learning to mitigate degradation effects, their frame-wise application to underwater 3D scenes does not explicitly enforce multi-view consistency.

NeRF~\cite{nerf} has revolutionized 3D scene representation, enabling high-quality rendering from multi-view images.
Recent underwater 3D reconstruction methods have sought to integrate the physics-based underwater imaging models into the NeRF framework~\cite{seathrunerf, beyonduw, neuraluw}. 
However, these approaches encounter several fundamental limitations in underwater scattering scenarios, such as artifacts in volumetric rendering, expensive ray sampling procedures, and slow convergence in underwater scenes due to the additional modeling complexity of water medium effects.
To overcome these limitations, recent efforts have transitioned from NeRF to 3DGS~\cite{3dgs}, which offers superior rendering efficiency.
Several underwater 3DGS variants ~\cite{watersplatting, seasplat, aquaticgs,restorgs} further introduce auxiliary neural modules or volumetric components to account for water-related degradation.
However, these hybrid designs also weaken the simplicity and efficiency advantages of 3DGS, while the quality of both their reconstruction and restoration remains limited in real-world underwater scenes.

To this end, we present WaterClear-GS, a novel underwater 3DGS framework built on an explicit optical proxy formulation.
Instead of reconstructing the full underwater medium, our method augments Gaussian primitives with physically motivated optical proxies that approximate dominant attenuation and scattering effects in a rendering-oriented manner, without introducing an additional medium network.
We further propose a dual-branch optimization strategy that jointly models the underwater observation branch and a latent clean branch, encouraging a practical separation between scene appearance and degradation while preserving geometric consistency.
As a result, WaterClear-GS provides a practical trade-off between simplicity, efficiency, and effectiveness, enabling high-quality underwater reconstruction and restoration within a unified 3DGS framework.

The main contributions of our work are as follows: 
\begin{itemize}
    \item We propose WaterClear-GS, a simple and explicit underwater 3DGS framework that augments Gaussian primitives with physically motivated local optical proxies, enabling degradation-aware scene representation without requiring an external medium network.
    \item We develop a dual-branch framework that jointly optimizes underwater scene reconstruction and learns a restoration-oriented latent clean appearance, supported by geometry and optical regularization to encourage stable geometry and plausible restored appearance.
    \item We collect a new underwater dataset, Water3D, featuring complex scattering conditions and diverse structural scenes to provide a more challenging test case.
    \item Extensive experiments show that WaterClear-GS achieves competitive performance on both novel view synthesis and underwater image restoration while maintaining 160+ FPS real-time rendering efficiency.
\end{itemize}
\section{Related Work}
\label{sec:related_work}

\subsection{Novel View Synthesis}
NeRF~\cite{nerf} represents scenes as continuous volumetric functions and enables high-quality novel-view synthesis from posed images.
Subsequent variants improve efficiency, quality, and generalization through explicit or factorized representations~\cite{ngp,tensorf,plenoxels,mipnerf,nerfactor,kplane}, while others address dynamic~\cite{dnerf,hypernerf,dynamicnerf} and large-scale scenes~\cite{blocknerf,meganerf}.
More recently, 3D Gaussian Splatting (3DGS)~\cite{3dgs} has achieved real-time rendering using explicit Gaussian primitives and alpha compositing.
Its extensions target sparse-view reconstruction~\cite{dngaussian,mvsplat,nerf-gs,slgaussian}, dynamic and large-scale scenes~\cite{gaussianflow,3dgstream,vastgs,citygs}, appearance modeling~\cite{relightgs,gaussianshader,llgaussian}, and compression~\cite{hac++,tntgs}.
These advances establish efficient and expressive scene representations, but they do not explicitly account for underwater image formation.

\subsection{Underwater Reconstruction and Restoration}
Underwater imaging suffers from severe degradation caused by wavelength-dependent attenuation and scattering.
Traditional single-image restoration methods~\cite{whitebalance,initial,underwater} can improve visual quality, but their frame-wise processing does not explicitly enforce consistency across views.
Recent neural rendering methods incorporate underwater image formation cues into NeRF- or 3DGS-based frameworks for multi-view reconstruction~\cite{neuraluw,beyonduw,waternerf,waterhenerf,seathrunerf,watersplatting,seasplat,aquaticgs,restorgs,uwgs,rusplatting}.
NeRF-based methods, such as WaterNeRF~\cite{waternerf}, WaterHE-NeRF~\cite{waterhenerf}, and SeaThru-NeRF~\cite{seathrunerf}, typically model underwater degradation by introducing medium-related components into volumetric rendering, but generally suffer from high training cost and slow rendering.
Underwater 3DGS approaches improve efficiency~\cite{uwgs,rusplatting,watersplatting,seasplat,aquaticgs,restorgs}, but some introduce additional medium representations: WaterSplatting~\cite{watersplatting} uses a volumetric water field, Aquatic-GS~\cite{aquaticgs} uses a neural water field, and SeaSplat~\cite{seasplat} optimizes learnable medium parameters.
Most closely related, 3D-UIR~\cite{3duir} decouples appearance and medium effects using explicit medium embeddings and pseudo-depth supervision.
In contrast, our method augments Gaussian primitives with local optical proxy parameters to approximate the dominant effects of underwater degradation in a fully explicit and efficient 3DGS framework.

\section{Methodology}

\begin{figure*}[ht]
\centering
\includegraphics[width=0.9\textwidth]{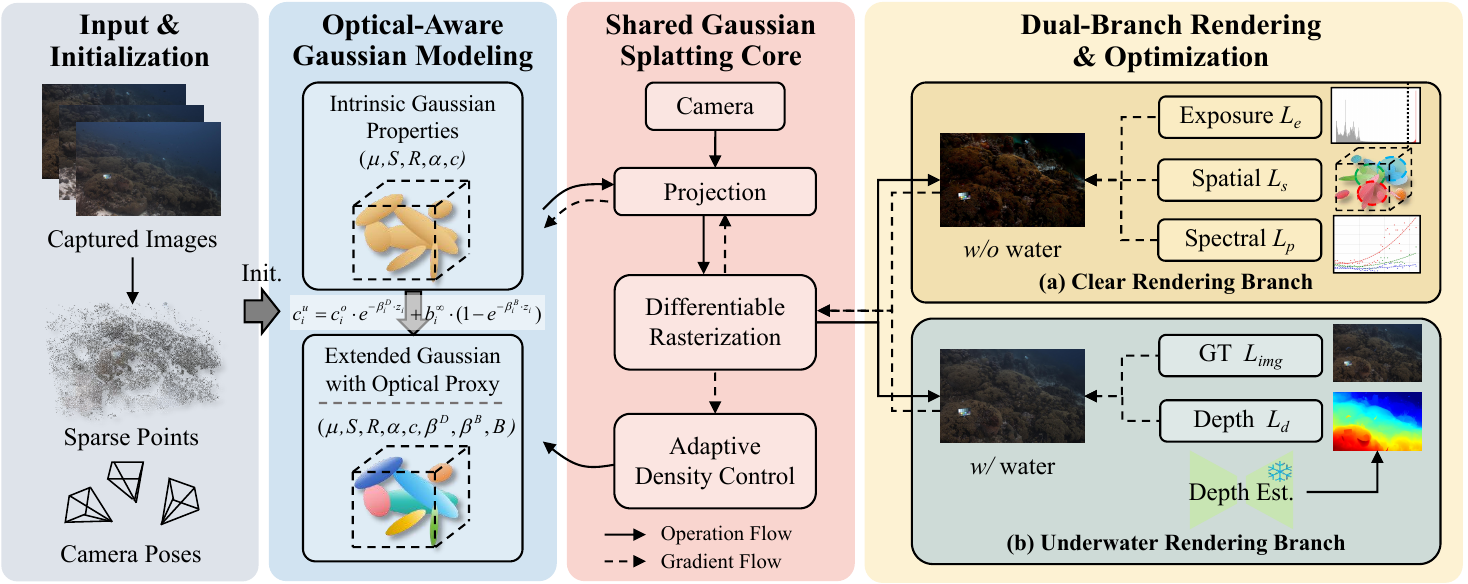} 
\caption{WaterClear-GS pipeline. Our method augments each Gaussian with wavelength-aware optical proxy parameters for attenuation, backscatter, and veiling light. A shared Gaussian splatting core supports dual-branch rendering: (a) the underwater branch fits the input observations, and (b) the clear branch learns a restoration-oriented latent clean appearance. Depth-guided supervision stabilizes geometry, and exposure, spatial, and spectral regularizations further improve restoration quality.}
\label{fig:pipeline}
\end{figure*}

\subsection{Preliminaries}

\textbf{3D Gaussian Splatting.} 
3DGS~\cite{3dgs} represents a scene with anisotropic Gaussian primitives parameterized by center $\mu$, covariance $\Sigma$, opacity $\alpha$, and SH appearance coefficients.
During rendering, visible Gaussians are projected to 2D ellipses and composited in depth order:
\begin{equation}
    C=\sum_{i=1}^{N} c_i \alpha_i \prod_{j<i}(1-\alpha_j),
\end{equation}
where $c_i$ is the color of the $i$-th Gaussian and $N$ is the number of contributing primitives.
Depth is rendered with the same transmittance weights by replacing $c_i$ with camera-space depth $z_i$.
Gaussian attributes and the primitive set are jointly optimized through photometric supervision and adaptive density control.

\vspace{1mm}
\noindent\textbf{Underwater Image Formation Model.}
Image formation in scattering media is affected in two ways.
On one hand, light reflected from objects undergoes distance- and wavelength-dependent attenuation as it propagates through the medium. 
On the other hand, in-scattering by suspended particles introduces a view-dependent backscatter component along the line of sight (LOS).
This backscatter is independent of scene content, accumulating with distance and causing reduced visibility, contrast, and severe color distortion.
Following the revised underwater model of Akkaynak and Treibitz~\cite{revised}, the image $I$ at distance $z$ is expressed as:
\begin{equation}
    I =J \,\cdot\, e^{-\beta^{D}(\mathbf v_{D})\,z}
    +
   B^{\infty}\,\cdot\,\bigl(1 - e^{-\beta^{B}(\mathbf v_{B})\,z}\bigr),
\end{equation}
where $J$ is the true radiance of the object, and $\beta^{D}(\mathbf v_{D})$ and $\beta^{B}(\mathbf v_{B})$ denote the wavelength-dependent attenuation and backscatter coefficients.
The vectors $\mathbf v_{D}$ and $\mathbf v_{B}$ encode the dependence on wavelength-related factors, including object reflectance, ambient light, water scattering, and the camera’s spectral response.
Both terms are channel-dependent, explaining the nonlinear interaction between distance, wavelength, and color distortion in underwater images.
The backscatter term $B^{\infty}$ represents veiling light (ambient light), typically non-uniform due to sunlight directionality and other factors~\cite{unveiling}.

\subsection{WaterClear-GS Overview} 
WaterClear-GS extends 3DGS with an optical-aware formulation tailored for underwater imaging.
As shown in Figure~\ref{fig:pipeline}, each Gaussian is augmented with wavelength-dependent optical proxy parameters for attenuation, backscatter, and veiling light, allowing the representation to capture local underwater degradation.
To jointly support underwater reconstruction and restoration-oriented appearance recovery, we use a dual-branch rendering strategy that produces both underwater and clean-appearance views from the same Gaussians, encouraging a practical separation between appearance and degradation while preserving geometric consistency.
We next describe the optical-aware Gaussian formulation, the dual-branch optimization strategy, and the associated regularizations.

\begin{figure}[t]
\centering
\includegraphics[width=\columnwidth]{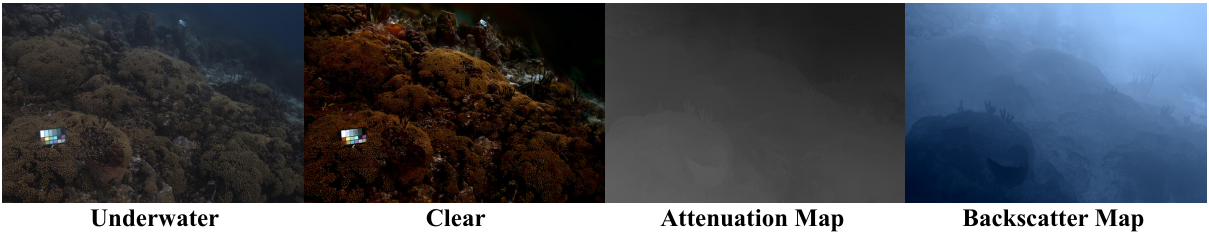}
\caption{Visualization of the OAGM decomposition. The underwater observation is represented using latent clear radiance together with direct-attenuation and backscatter proxies. These components are rendering-oriented variables rather than calibrated medium measurements.}
\label{fig:decom}
\end{figure}

\vspace{1mm}
\noindent\textbf{Optical-Aware Gaussian Modeling (OAGM).}
Instead of introducing separate medium representations, we embed optical parameters directly into Gaussian primitives.
This casts degradation modeling as part of explicit representation learning rather than a separate medium-estimation network.
We assign each Gaussian local wavelength-dependent optical proxy parameters that modulate its color before rasterization.
For the $i$-th Gaussian, let ${c}_i^o \in \mathbb{R}^3$ denote its base RGB color, and let $d_i = (T_{wc}\tilde{\mu}_i)_z$ denote the camera-space depth of its homogeneous center $\tilde{\mu}_i$ under the world-to-camera transform $T_{wc}$.
We define channel-wise attenuation, backscatter, and veiling-light proxies ${\beta}^D_{i}, {\beta}^B_{i}, {B}_i \in \mathbb{R}^3$, and compute the underwater color as:
\begin{equation}
    {c}_i^{u} 
    = {T}^D_{i} \odot {c}_i^{o}
    + (1 - {T}^B_{i}) \odot {B}_i,
    \label{eq:underwater_color}
\end{equation}
where ${T}^D_i = \exp(-{\beta}^D_i d_i)$ and ${T}^B_i = \exp(-{\beta}^B_i d_i)$ are wavelength-dependent transmittances applied element-wise over RGB.
During $\alpha$-blending, ${c}_i^{u}$ replaces the original color ${c}_i^{o}$, yielding the rendered pixel:
\begin{equation}
    {C}^u = \sum_{i=1}^{N} {c}_i^u \,\alpha_i 
    \prod_{j<i} (1 - \alpha_j).
\end{equation}
The optical proxy parameters ${\beta}^D_i, {\beta}^B_i, {B}_i$ are fully differentiable and optimized jointly with all other Gaussian attributes, enabling the model to capture spatially varying underwater degradation cues.
They are effective rendering-oriented variables, not calibrated medium coefficients.
Thus, OAGM is a first-order approximation of direct attenuation and single-backscatter effects rather than a full radiative-transfer simulation.
Figure~\ref{fig:decom} visualizes how the latent clear radiance and the learned degradation proxies contribute to the underwater observation.
OAGM does not introduce opacity as a new attribute: opacity retains its standard 3DGS role in visibility and compositing, while the added proxies modulate appearance before rasterization.

\vspace{1mm}
\noindent\textbf{Dual-Branch Optimization Strategy.}
To support both reconstruction and restoration, we optimize the scene under two rendering conditions: \textit{underwater} and \textit{clear}.
The \textit{underwater} branch renders images using the full underwater formation model. 
By fitting the input underwater images, this branch provides geometric supervision and enforces observation consistency.
Instead of treating restoration as a post-processing task, we jointly learn a clean latent appearance within the same Gaussian representation.
The \textit{clear} branch renders the same Gaussian primitives after removing the learned underwater degradation terms, yielding a latent clean-appearance rendering.
This branch learns a restoration-oriented latent clean appearance without direct ground-truth supervision.
The decomposition is therefore not strictly identifiable.
Rather, the shared geometry, underwater reconstruction branch, and restoration regularizers discourage degenerate solutions in which scattering-like effects are absorbed into surface appearance.
Since both branches share the same Gaussians, they provide complementary constraints for a stable practical separation of appearance and degradation.

\subsection{Optimization for Reconstruction}

\textbf{Depth-Guided Geometric Regularization.}
Depth is particularly important in underwater scenes because attenuation and scattering depend strongly on propagation distance, while weak textures often cause floating artifacts and geometric instability.
To alleviate this issue, we use a depth-guided regularization term to stabilize geometry learning.
Following prior methods~\cite{rusplatting,dehazegs}, we adopt DepthAnythingV2~\cite{depthanythingv2} to provide pseudo-depth supervision.
Typical pixel-wise L1 and L2 losses on depth maps are sensitive to scale and shift ambiguities.
To emphasize structural consistency rather than absolute depth values, we employ a loss based on the Pearson correlation coefficient (PCC) to maximize the similarity between the predicted depth $D_{pred}$ and the pseudo-depth $D_{pseudo}$:
\begin{equation}
L_{\text{d}} = 1 - 
\frac{\mathrm{Cov}(D_{pred}, D_{pseudo})}
{\sigma(D_{pred})\,\sigma(D_{pseudo})}.
\label{equ:pcc}
\end{equation}
This correlation-based loss focuses on relative depth structure rather than absolute metric values, making it more robust to scale and shift errors in monocular predictions.

\vspace{1mm}
\noindent\textbf{Perception-Driven Image Loss.}
Underwater images often exhibit unbalanced illumination and color-dependent degradation, while conventional image losses treat all pixels equally and therefore under-supervise dark regions.
Inspired by RawNeRF~\cite{rawnerf}, we perform a perception-driven transformation $\psi(\cdot)$ that adaptively rescales image intensities to balance gradient magnitudes across illumination levels.
We define a gradient-rescaling mapping for the estimated image $\hat{y}$ and the ground-truth image $y$:
\begin{equation}
\psi(y)=\frac{y}{\operatorname{sg}(\hat{y})+\epsilon},
\label{eq:phi_map}
\end{equation}
where $\epsilon{=}10^{-3}$ ensures numerical stability and $\operatorname{sg}(\cdot)$ denotes the stop-gradient operator.
Using $\operatorname{sg}(\hat{y})$ allows the model to adaptively emphasize regions according to its current predictions.
This transformation behaves like a differentiable inverse-intensity tone mapping, increasing supervision in dark regions while remaining stable in bright areas.

The photometric reconstruction loss is evaluated in the perceptual domain:
\begin{equation}
    \mathcal{L}_{\text{W-L2}}
=\frac{1}{N}\lVert \psi(\hat{y})-\psi(y) \rVert_2^2,
\end{equation}
where $N$ denotes the total number of pixels.
Similarly, the structural similarity loss is computed between the transformed images:
\begin{equation}
\mathcal{L}_{\text{W-DSSIM}}
=1-\operatorname{SSIM}\!\big(\psi(\hat{y}),\,\psi(y)\big),
\end{equation}
which effectively measures perceptual similarity in a tone-mapped domain, enhancing sensitivity to structural errors in low-intensity regions.
Then the final image reconstruction loss combines the two terms:
\begin{equation}
\mathcal{L}_{\text{img}}
=(1-\lambda_{\text{SSIM}})\,\mathcal{L}_{\text{W-L2}}
+\lambda_{\text{SSIM}}\,\mathcal{L}_{\text{W-DSSIM}}.
\label{equ:img}
\end{equation}

\subsection{Optimization for Restoration}
Without ground-truth supervision, the latent clean appearance and the estimated degradation can be ambiguously decomposed.
To reduce this ambiguity, we employ three complementary components: an exposure constraint, a spatially adaptive regularization, and a physics-informed spectral prior.

\vspace{1mm}
\noindent\textbf{Exposure Constraint.}
Underwater scenes often contain uneven illumination, and the interplay between attenuation estimation and color recovery can make the restored results susceptible to overexposure.
This issue can be further amplified by the perception-driven reconstruction loss, which up-weights dark regions and may inadvertently encourage the network to brighten the output to reduce the weighted error.
To maintain a visually natural dynamic range consistent with human perception, we introduce an exposure regularization term that penalizes over-bright regions in the restored image:
\begin{equation}
\begin{aligned}
\mathcal{L}_{e}
&= \frac{1}{3|\Omega|}\sum_{p \in \Omega}\sum_{c \in \{r,g,b\}}
\max\!\left(I_{\mathrm{clear}}^{c}(p) - \tau, 0\right),
\end{aligned}
\label{equ:expo}
\end{equation}
where $I_{\mathrm{clear}}^{c}(p)$ denotes channel $c$ of the latent clean-appearance rendering at pixel $p$, $\Omega$ is the set of image pixels, and $\tau$ is the brightness threshold.
This constraint helps suppress overexposure and encourages balanced illumination across the restored scene.

\vspace{1mm}
\noindent\textbf{Spatially Adaptive Regularization}.
To model spatially varying underwater degradation, we impose a spatially adaptive regularization based on local consistency.
It applies distance-weighted smoothing to the optical parameters (${\beta^D}$, ${\beta^B}$, $B$).
Specifically, for each visible Gaussian from the current viewpoint, we perform local smoothing within a spherical neighborhood of radius $r$, constraining parameter consistency with neighboring Gaussians through the following loss function:
\begin{equation}
L_{s} = \frac{1}{|V|} \sum_{i \in V} \frac{\sum_{j \in \mathcal{N}_i} \| \theta_i - \theta_j \|_2^2 \cdot w_{ij}}{\sum_{j \in \mathcal{N}_i} w_{ij}},
\label{equ:smooth}
\end{equation}
where $V$ represents the set of valid Gaussians with sufficient neighbors ($\geq n_{min}$), $\mathcal{N}_i$ denotes the neighborhood of Gaussian $i$, $\theta$ represents the water optical parameters to be smoothed, and $w_{ij} = 1/(d_{ij} + \varepsilon)$ is a weight based on the Euclidean distance $d_{ij}$.
This local smoothing encourages neighborhood consistency without enforcing global uniformity, allowing optical parameters to vary with depth and local water conditions.

\vspace{1mm}
\noindent\textbf{Physics-Informed Spectral Regularization}.
To encourage plausible spectral behavior, we introduce a lightweight regularizer that softly constrains RGB ordering among the per-Gaussian optical proxies.
For each Gaussian, attenuation, backscatter, and veiling light are defined separately for the RGB channels: ${\beta}^D(i)=[\beta^D_r, \beta^D_g, \beta^D_b]_i$, ${\beta}^B(i)=[\beta^B_r, \beta^B_g, \beta^B_b]_i$, and ${B}(i)=[B_r, B_g, B_b]_i$.
We define their combined spectral differences as:
\begin{equation}
  {\Delta}_i = 
  [\, \Delta {\beta}^D(i),~
     \Delta {\beta}^B(i),~
     \Delta {B}(i) \,],
\end{equation}  
\begin{equation}
\left\{
\begin{aligned}
  \Delta{\beta}^D(i) &= [\,\beta^D_g(i)-\beta^D_r(i),~ \beta^D_b(i)-\beta^D_g(i)\,],\\
  \Delta{\beta}^B(i) &= [\,\beta^B_r(i)-\beta^B_g(i),~ \beta^B_g(i)-\beta^B_b(i)\,],\\
  \Delta{B}(i) &= [\,B_r(i)-B_g(i),~ B_g(i)-B_b(i)\,].
\end{aligned}
\right.
\label{eq:delta_definitions}
\end{equation}

The physical prior loss softly penalizes violations of the expected spectral ordering using a \emph{Softplus} function:
\begin{equation}
  L_{p}
  = \mathbb{E}_i \big[
      \text{Softplus}(\Delta_i + \delta)
    \big],
  \quad
\label{equ:phy}
\end{equation}
where $\delta$ is a small tolerance margin.
It encourages common underwater spectral trends, such as stronger red attenuation, stronger blue scattering, and bluish veiling light~\cite{revised,unveiling}, while remaining differentiable for stable optimization.
This is a weak prior, not a universal physical ordering: its margin can be relaxed for different water conditions.

The overall objective of WaterClear-GS is defined as:
\begin{equation}
L_{total} = L_{img} + \lambda_{d}L_{d} + \lambda_{s}L_{s} + \lambda_{p}L_{p} + \lambda_{e}L_{e}.
\end{equation}

\section{Experiments}

\subsection{Experimental Setup}

\begin{table*}[t]
\centering
\small
\caption{Average NVS results on SeaThru-NeRF~\cite{seathrunerf}, Submerged3D~\cite{rusplatting}, and Water3D. Time and FPS are averaged across all scenes. Top-three results are marked by \colorbox{red!25}{first}, \colorbox{orange!25}{second}, and \colorbox{yellow!25}{third}.}
\resizebox{0.9\textwidth}{!}{%
\begin{tabular}{l|ccc|ccc|ccc|c|c}
\toprule
\multirow{2}{*}{Method}
 & \multicolumn{3}{c|}{SeaThru-NeRF}
 & \multicolumn{3}{c|}{Submerged3D}
 & \multicolumn{3}{c|}{Water3D}
 & Avg.
 & Avg.\\
\cmidrule(lr){2-4}
\cmidrule(lr){5-7}
\cmidrule(lr){8-10}
 & PSNR$\uparrow$ & SSIM$\uparrow$ & LPIPS$\downarrow$
 & PSNR$\uparrow$ & SSIM$\uparrow$ & LPIPS$\downarrow$
 & PSNR$\uparrow$ & SSIM$\uparrow$ & LPIPS$\downarrow$
 & Time$\downarrow$
 & FPS$\uparrow$ \\
\midrule
SeaThru-NeRF~\cite{seathrunerf}
 & 26.47 & 0.819 & 0.282
 & 17.22 & 0.492 & 0.598
 & 24.37 & 0.722 & 0.460
 & $>$10h & $<$0.1 \\
3DGS~\cite{3dgs}
 & 26.71 & 0.901 & 0.204
 & 22.71 & 0.767 & 0.315
 & \cellcolor{yellow!25}29.19 & 0.858 & 0.307
 & \cellcolor{red!25}5.37min & \cellcolor{red!25}234.53 \\
UW-GS~\cite{uwgs}
 & 27.15 & 0.899 & 0.202
 & 21.98 & 0.731 & 0.356
 & 26.55 & 0.849 & 0.315
 & 30.79min & 31.72 \\
SeaSplat~\cite{seasplat}
 & 27.84 & 0.909 & 0.178
 & 23.14 & 0.769 & 0.319
 & 28.86 & \cellcolor{yellow!25}0.861 & 0.293
 & 29.38min & 60.28 \\
WaterSplatting~\cite{watersplatting}
 & \cellcolor{orange!25}29.57 & \cellcolor{orange!25}0.925 & \cellcolor{orange!25}0.150
 & \cellcolor{yellow!25}24.55 & 0.768 & 0.297
 & 29.14 & 0.853 & 0.303
 & \cellcolor{orange!25}10.11min & \cellcolor{yellow!25}75.10 \\
RUSplatting~\cite{rusplatting}
 & 28.53 & \cellcolor{yellow!25}0.917 & 0.163
 & 24.21 & \cellcolor{yellow!25}0.788 & \cellcolor{yellow!25}0.286
 & \cellcolor{orange!25}29.45 & \cellcolor{orange!25}0.864 & \cellcolor{yellow!25}0.291
 & 23.79min & 52.08 \\
3D-UIR~\cite{3duir}
 & \cellcolor{yellow!25}29.06 & 0.914 & \cellcolor{yellow!25}0.161
 & \cellcolor{orange!25}25.23 & \cellcolor{orange!25}0.805 & \cellcolor{orange!25}0.277
 & \cellcolor{orange!25}29.45 & \cellcolor{yellow!25}0.861 & \cellcolor{orange!25}0.290
 & 56.50min & 70.58 \\
\textbf{Ours}
 & \cellcolor{red!25}29.93 & \cellcolor{red!25}0.931 & \cellcolor{red!25}0.149
 & \cellcolor{red!25}25.40 & \cellcolor{red!25}0.812 & \cellcolor{red!25}0.270
 & \cellcolor{red!25}29.62 & \cellcolor{red!25}0.869 & \cellcolor{red!25}0.283
 & \cellcolor{yellow!25}14.25min & \cellcolor{orange!25}175.23 \\
\bottomrule
\end{tabular}}
\label{tab:NVS}
\end{table*}

\begin{figure*}[!t]
\centering
\includegraphics[width=\textwidth]{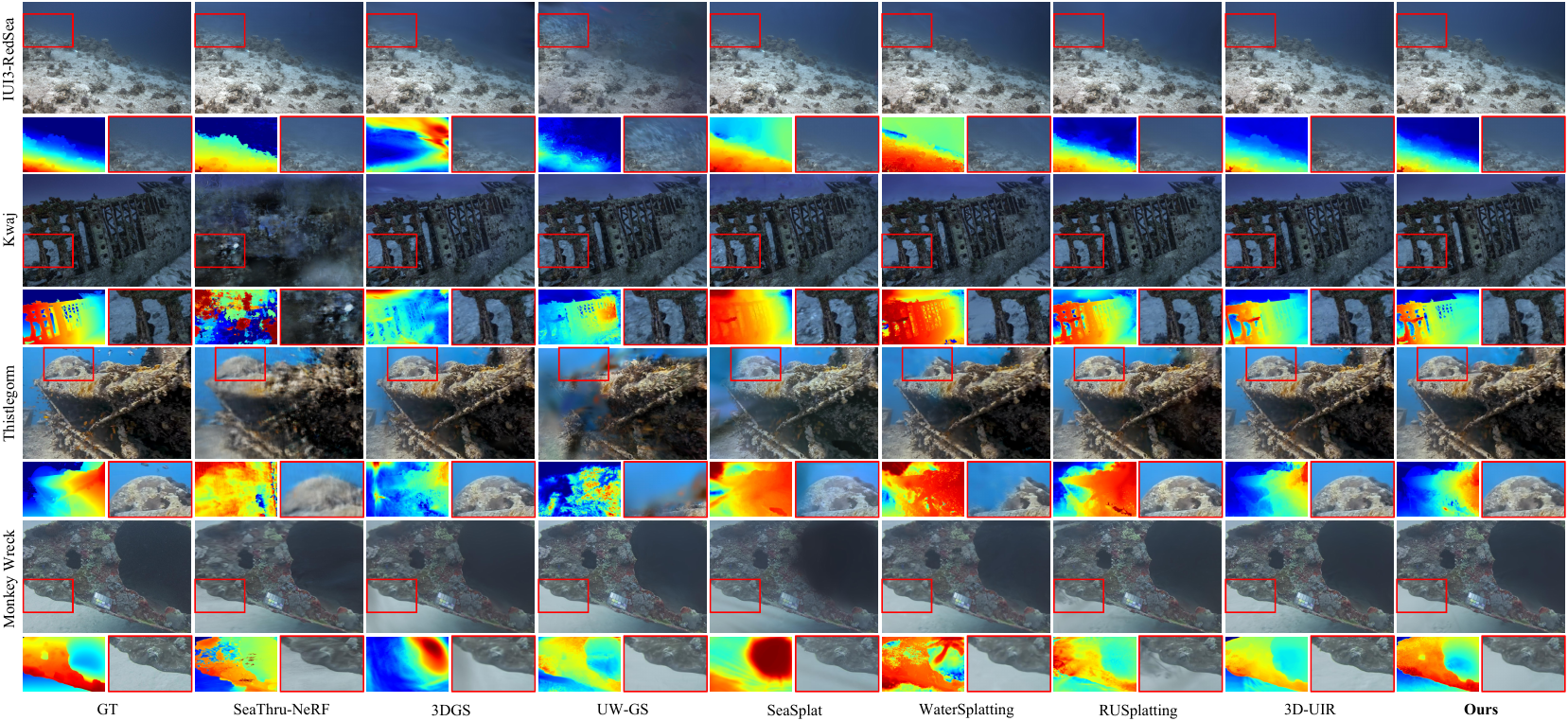} 
\caption{Qualitative results of the NVS task.
We present rendered underwater images and their depth maps. The pseudo-depth maps are used as a reference. Zoomed-in regions (highlighted with red bounding boxes) illustrate detailed differences. Our method consistently preserves geometric structures and fine-level details in most cases.}
\label{fig:NVS}
\end{figure*}

\textbf{Public Datasets.}
We evaluate on SeaThru-NeRF~\cite{seathrunerf} and Submerged3D~\cite{rusplatting}, which contain four coral-reef and four deep-sea shipwreck scenes, respectively.

\vspace{1mm}
\noindent\textbf{Water3D Dataset.}
To broaden scene content and viewpoint coverage, we construct Water3D with four real-world scenes.
The Aqaba and Thistlegorm scenes are collected from online shipwreck videos under different water conditions, while Monkey Wreck and Frame are captured by us in the Sabang Sea using a DJI Action 4 camera.
Standard color charts are placed in the two self-captured scenes for color evaluation.
We follow the 3DGS train/test protocol and estimate camera parameters using COLMAP~\cite{colmap}.

\vspace{1mm}
\noindent\textbf{Baseline Methods.}
We compare our method with vanilla 3DGS~\cite{3dgs} and representative underwater methods:
SeaThru-NeRF~\cite{seathrunerf}, UW-GS~\cite{uwgs}, WaterSplatting~\cite{watersplatting}, SeaSplat~\cite{seasplat}, RUSplatting~\cite{rusplatting}, and the recent 3D-UIR~\cite{3duir}.
\vspace{1mm}
\noindent\textbf{Metrics.}
For the NVS task, we evaluate PSNR, SSIM~\cite{ssim}, and LPIPS~\cite{lpips} on the test sets.
We also report the training time and rendering FPS to evaluate model efficiency.
Following prior underwater enhancement studies~\cite{unveiling, domain, ushape}, we adopt the CIEDE2000 color difference ($\Delta E_{00}$)~\cite{ciede2000} and the average angular error $\bar{\psi}$ (in degrees)~\cite{sucre} to assess color restoration quality on datasets containing standard color charts.
To complement these reference-based color-chart metrics, we report NIQE~\cite{niqe}, a no-reference measure of natural-image statistics for which lower is better.
We also use UIQM~\cite{uiqm} and UCIQE~\cite{uciqe}, two underwater-specific no-reference metrics that evaluate perceptual quality from colorfulness, sharpness, and contrast statistics, for which higher is better.
Because no-reference underwater metrics may favor excessive colorfulness or contrast, we interpret them jointly with the color-chart metrics and qualitative comparisons.

\vspace{1mm}
\noindent\textbf{Implementation Details.}
Our implementation builds on 3DGS~\cite{3dgs} and uses RGB values without additional SH coefficients.
All scenes are optimized for 15,000 iterations, with densification every 500 iterations, on a single NVIDIA RTX 4090 GPU.

\begin{figure*}[!t]
\centering
\includegraphics[width=\textwidth]{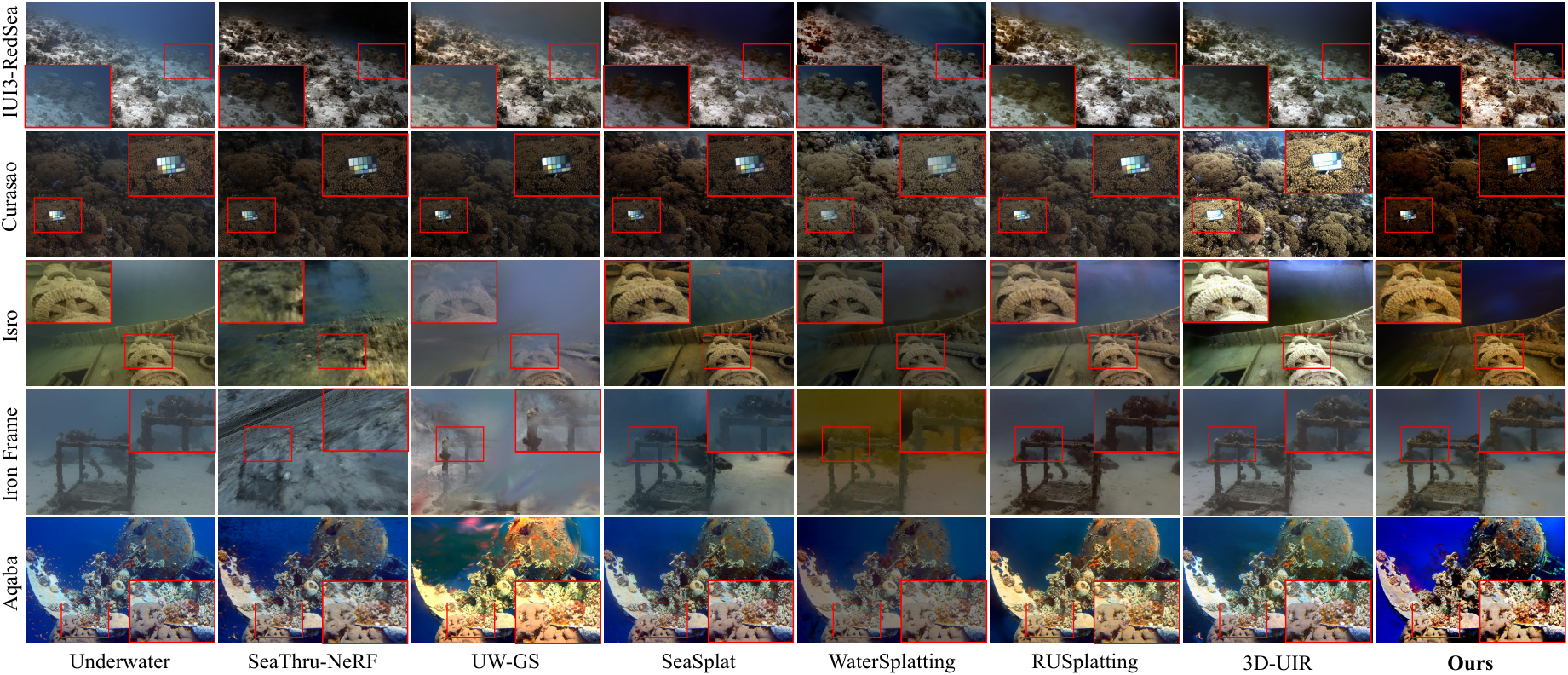}
\caption{Qualitative comparison for the UIR task.
Across the shown scenes, WaterClear-GS generally produces more balanced colors and clearer local details, while some baselines exhibit underexposure, residual haze, or color shifts.}
\label{fig:UIR-all}
\end{figure*}

\subsection{Novel View Synthesis Results}
Table~\ref{tab:NVS} reports dataset-level average results for the NVS task.
Our method achieves the best average PSNR, SSIM, and LPIPS on all three datasets.
It also consistently outperforms UW-GS and RUSplatting, despite their use of depth-related priors or pseudo-depth supervision.
Compared with the recent 3D-UIR, WaterClear-GS improves PSNR, SSIM, and LPIPS across all three datasets while training about four times faster and rendering more than twice as fast.
WaterClear-GS renders at 175.23 FPS on average, approximately \(2\times\) faster than WaterSplatting and orders of magnitude faster than the NeRF-based SeaThru-NeRF, by embedding underwater optical effects directly into Gaussian rendering without auxiliary volumetric fields or MLP estimators.
Vanilla 3DGS remains faster because it omits underwater degradation modeling; among the evaluated degradation-aware methods, WaterClear-GS provides the highest rendering FPS.
Figure~\ref{fig:NVS} further shows cleaner structures, sharper details, and fewer floating artifacts across diverse underwater scenes, particularly under strong degradation.

\begin{table}[t]
\centering
\small
\caption{Average UIR quality comparison.}
\setlength{\tabcolsep}{2pt}
\begin{tabular}{l|cc|ccc}
\toprule
Method & $\Delta E_{00}\downarrow$ & $\bar{\psi}\downarrow$ & NIQE$\downarrow$ & UIQM$\uparrow$ & UCIQE$\uparrow$ \\
\midrule
SeaThru-NeRF~\cite{seathrunerf} & 24.73 & 24.82 & 5.31 & 2.91 & 0.26 \\
UW-GS~\cite{uwgs} & \cellcolor{orange!25}19.13 & \cellcolor{orange!25}21.44 & 4.11 & \cellcolor{orange!25}4.16 & 0.28 \\
SeaSplat~\cite{seasplat} & \cellcolor{yellow!25}20.55 & \cellcolor{yellow!25}21.68 & 4.13 & 3.65 & \cellcolor{yellow!25}0.34 \\
WaterSplatting~\cite{watersplatting} & 24.45 & 23.99 & 4.75 & \cellcolor{yellow!25}4.08 & 0.24 \\
RUSplatting~\cite{rusplatting} & 22.44 & 22.46 & \cellcolor{red!25}3.99 & 4.02 & 0.30 \\
3D-UIR~\cite{3duir} & 22.37 & 23.40 & \cellcolor{yellow!25}4.08 & 3.79 & \cellcolor{orange!25}0.35 \\
\textbf{Ours} & \cellcolor{red!25}19.09 & \cellcolor{red!25}20.24 & \cellcolor{orange!25}4.02 & \cellcolor{red!25}4.19 & \cellcolor{red!25}0.39 \\
\bottomrule
\end{tabular}
\label{tab:UIR}
\end{table}

\subsection{Underwater Image Restoration Results}
Figure~\ref{fig:UIR-all} presents the qualitative comparison of underwater image restoration.
SeaThru-NeRF often fails in heavily degraded scenes, UW-GS introduces noticeable color shifts in Aqaba, and WaterSplatting produces darker, hazier results across the evaluated datasets.
Compared with SeaSplat and RUSplatting, WaterClear-GS generally yields more plausible restorations with better color balance and sharper fine-scale structures.
To quantitatively evaluate restoration quality, Table~\ref{tab:UIR} reports the average color-chart metrics $\Delta E_{00}$ and $\bar{\psi}$ together with the no-reference metrics NIQE, UIQM, and UCIQE.
WaterClear-GS obtains the lowest average $\Delta E_{00}$ and $\bar{\psi}$, the best UIQM and UCIQE, and the second-best NIQE, demonstrating strong color fidelity and overall perceptual quality.

\subsection{Model Analysis}

\begin{figure}[t]
\centering
\includegraphics[width=0.85\columnwidth]{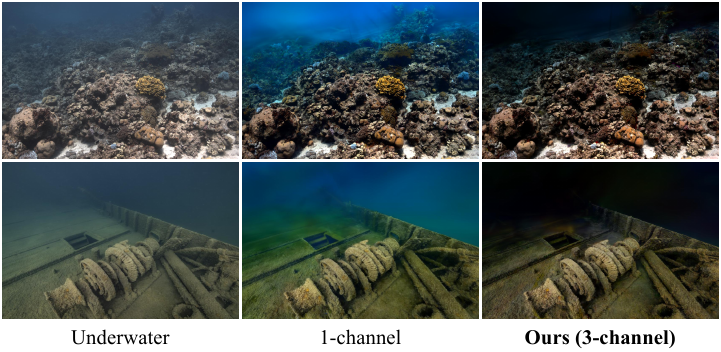} 
\caption{Results of single-channel and RGB optical modeling.}
\label{fig:AB-channel}
\end{figure}

\textbf{Effect of Multichannel Optical Parameters.}
To validate the necessity of wavelength-dependent optical modeling, we replace the three-channel optical parameters ${\beta}^D, {\beta}^B, {B} \in \mathbb{R}^3$ with single-channel scalars shared across RGB. 
As shown in Figure~\ref{fig:AB-channel}, single-channel modeling causes noticeable hue shifts and unrealistic bluish artifacts in distant regions, indicating that it fails to capture the spectral behavior of underwater degradation.
This result shows that RGB-channel optical modeling is important for plausible color recovery.
\begin{figure*}[!t]
\centering
\includegraphics[width=0.85\textwidth]{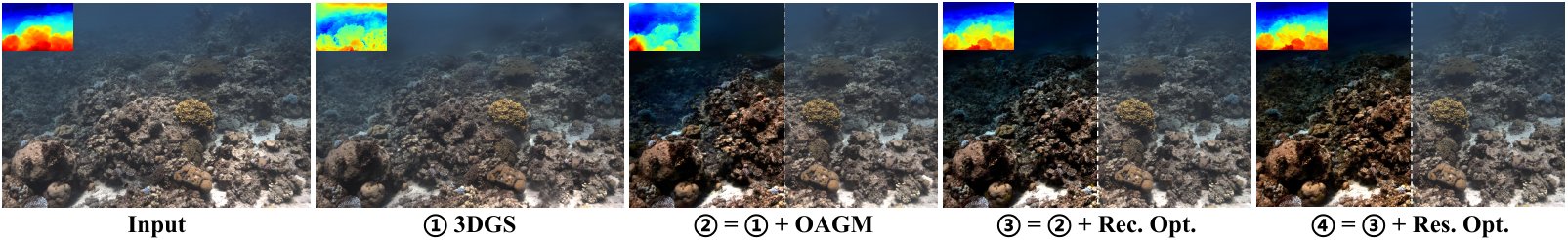}
\caption{Module-level visual ablation. OAGM enables the latent clean branch, reconstruction optimization improves structural fidelity, and restoration optimization further refines the recovered appearance.}
\label{fig:ab}
\end{figure*}

\vspace{1mm}
\noindent\textbf{Controlled Synthetic Validation.}
\begin{figure}[t]
\centering
\includegraphics[width=\columnwidth]{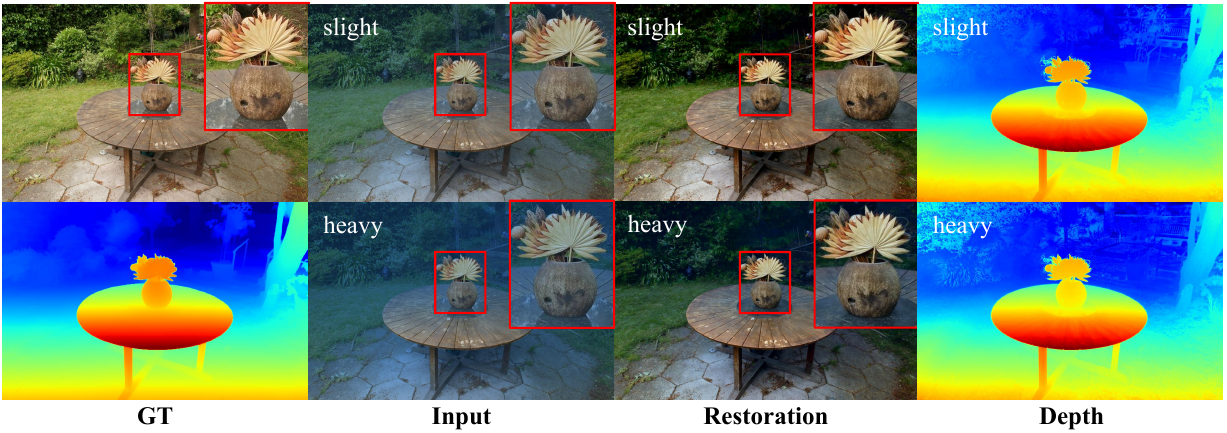}
\caption{Restoration under controlled synthetic turbidity. The model is optimized from degraded observations without using the clean reference as supervision.}
\label{fig:syn}
\end{figure}
We additionally evaluate the unsupervised clean branch under controlled synthetic turbidity levels, retaining the clean image only as a visual reference.
As shown in Figure~\ref{fig:syn}, WaterClear-GS recovers a stable latent clean appearance under both mild and severe degradation.
This result supports that the decomposition is driven by the shared rendering model and regularization rather than paired clean supervision, while it does not imply strict physical identifiability of the learned proxies.

\vspace{1mm}
\noindent\textbf{Impact of Color Representation.}
Since scattering suppresses high-frequency view-dependent effects, degree-0 SHs provide a favorable quality-efficiency trade-off.

\subsection{Ablation Study}

\begin{table}[t]
\centering
\setlength{\tabcolsep}{2pt}
\caption{Ablation study of the main components on the SeaThru-NeRF dataset. Base denotes our 3DGS backbone trained with the same hyperparameters as WaterClear-GS.}
\resizebox{0.95\linewidth}{!}{%
\begin{tabular}{c c c c|c c c}
\toprule
Base & OAGM & Rec. Opt. & Res. Opt. & PSNR$\uparrow$ & SSIM$\uparrow$ & LPIPS$\downarrow$ \\
\midrule
\ding{51}  &\ding{55}  &\ding{55}  &\ding{55}  & 28.57 & 0.902 & 0.235 \\
\ding{51}  &\ding{55}  & \ding{51}  & \ding{55} & 29.10 & 0.924 & 0.161 \\
\ding{51}  &\ding{51}  & \ding{55} & \ding{55} & 28.71 & 0.906 & 0.224 \\
\ding{51}  &\ding{51}  & \ding{55} & \ding{51}  & 28.83 & 0.905 & 0.224 \\
\ding{51}  &\ding{51}  & \ding{51}  & \ding{55} & \underline{29.79} & \underline{0.929} & \underline{0.151} \\
\ding{51}  &\ding{51}  & \ding{51}  & \ding{51}  & \textbf{29.93} & \textbf{0.931} & \textbf{0.149} \\
\bottomrule
\end{tabular}}
\label{tab:ablation}
\end{table}
\begin{table}[t]
\centering
\small
\setlength{\tabcolsep}{2pt}
\caption{Restoration-loss ablation on SeaThru-NeRF. $\Delta E_{00}$ and $\bar\psi$ measure latent clear-view color fidelity.}
\begin{tabular}{c|c c c|c c}
\toprule
Setting & PSNR$\uparrow$ & SSIM$\uparrow$ & LPIPS$\downarrow$ & $\Delta E_{00}\downarrow$ & $\bar\psi\downarrow$\\
\midrule
w/o $L_e$ & 29.80 & 0.930 & 0.150 & 16.26 & 15.05\\
w/o $L_s$ & \textbf{30.00} & 0.931 & 0.149 & 15.84 & 15.77\\
w/o $L_p$ & 29.97 & 0.930 & 0.151 & 16.49 & 15.97\\
Full model & 29.93 & \textbf{0.931} & \textbf{0.149} & \textbf{14.56} & \textbf{14.74}\\
\bottomrule
\end{tabular}
\label{tab:ab-restoration}
\end{table}

To validate the effectiveness of WaterClear-GS, we conduct ablation studies at two levels.
Table~\ref{tab:ablation} first evaluates the three major components of the framework, namely optical-aware Gaussian modeling (OAGM), reconstruction optimization, and restoration optimization.
Reconstruction optimization serves as the primary driver of reconstruction quality, while OAGM and restoration optimization are mainly designed for appearance modeling.
When introduced without reconstruction optimization, the added optical modeling and restoration objectives result in only marginal NVS gains.
When combined, the three components provide complementary benefits in both geometry and appearance, yielding the strongest NVS metrics in Table~\ref{tab:ablation}.
Table~\ref{tab:ab-restoration} further shows that each restoration loss improves color fidelity on the latent clear branch, although its effect on NVS metrics is limited.
Figure~\ref{fig:ab} provides the corresponding module-level visual evidence.
It shows that OAGM enables the latent clean branch, reconstruction optimization improves structural fidelity, and restoration optimization further refines the recovered appearance.

\section{Conclusion and Discussion}
We presented WaterClear-GS, an explicit optical-aware underwater 3DGS framework that unifies scene reconstruction and restoration while preserving real-time performance.
By augmenting Gaussian primitives with wavelength-aware optical proxies and a dual-branch optimization strategy, our method efficiently models dominant underwater degradation.
Experiments show strong performance on both novel view synthesis and underwater image restoration at real-time speed (160+ FPS).
Our current formulation mainly targets static scenes under stable imaging conditions.
In extremely turbid or poorly illuminated scenes, severe information loss can leave residual color casts and artifacts;
distant open-water regions and dynamic floating objects are likewise difficult to separate from scene appearance.
Improving robustness in such challenging scenarios while maintaining simplicity and efficiency remains an important direction for future work.

{
    \small
    \bibliographystyle{ieeenat_fullname}
    \bibliography{main}
}

\end{document}